\documentclass[journal,twoside]{IEEEtran}
\usepackage{amsmath,amsfonts}
\usepackage{algorithmic}
\usepackage{algorithm}
\usepackage{array}
\usepackage[caption=false,font=normalsize,labelfont=sf,textfont=sf]{subfig}
\usepackage{textcomp}
\usepackage{stfloats}
\usepackage{url}
\usepackage{verbatim}
\usepackage{graphicx}
\usepackage{cite}
\usepackage{graphicx}
\usepackage{xcolor}
\usepackage{booktabs}
\usepackage{multirow}

\newcommand{\fst}[1]{\textcolor{red}{#1}}
\newcommand{\snd}[1]{\textcolor{blue}{#1}}

\begin{document}

\title{Image-Condition-Controlled Diffusion Network with \\ Large-Scale CT Pretraining for CT-Free Attenuation Correction in Dynamic Cardiac PET}

\author{Zhejia Zhang, Scott M. Leonard, Jinkui Hao, Ryan J. Avery, Richard L. Weinberg, and Bo Zhou%
\thanks{This work has been submitted to the IEEE for possible publication. Copyright may be transferred without notice, after which this version may no longer be accessible. \textit{(Corresponding authors: Zhejia Zhang; Bo Zhou.)}}%
\thanks{This work involved human subjects in its research. Approval of all ethical and experimental procedures and protocols was granted by the Institutional Review Board of Northwestern University under Protocol No. STU00222257.}%
\thanks{Zhejia Zhang is with the Department of Radiology, Northwestern University, Chicago, IL 60611 USA, and also with the Department of Electrical and Computer Engineering, Northwestern University, Evanston, IL 60208 USA (e-mail: zhangzhejia2002@gmail.com).}%
\thanks{Scott M. Leonard, Jinkui Hao, Ryan J. Avery, and Bo Zhou are with the Department of Radiology, Northwestern University, Chicago, IL 60611 USA (e-mail: bo.zhou@northwestern.edu).}%
\thanks{Richard L. Weinberg is with the Division of Cardiology, Department of Medicine, Northwestern University, Chicago, IL 60611 USA.}}

\markboth{IEEE Transactions on Radiation and Plasma Medical Sciences}%
{Zhang \MakeLowercase{\textit{et al.}}: Controlled Diffusion with Large-Scale CT Pretraining for CT-Free Attenuation Correction}

\maketitle
\begin{abstract}
Positron emission tomography (PET) provides quantitative information on myocardial perfusion. In dynamic PET, tracer kinetics are tracked through a series of reconstructed time frames. Reliable attenuation correction is essential for accurate measurements. Clinical attenuation maps ($\mu$-maps) are typically derived from computed tomography (CT), which adds radiation exposure and may introduce PET--CT misregistration due to respiratory and cardiac motion. Moreover, supervised CT-free methods rely on paired non-attenuation-corrected (NAC) PET and CT data, which are scarce, whereas large-scale unpaired CT data are more readily available. To address these challenges, we propose Large-scale Unpaired CT-Informed Diffusion (LUCID), a framework that transfers anatomical priors learned from 10,000 chest CT volumes in the public CT-RATE dataset to $\mu$-map synthesis for dynamic cardiac $^{82}$Rb PET. First, a convolutional neural network (CNN) predicts an initial $\mu$-map from each NAC PET frame. Second, the CT-pretrained diffusion backbone refines a low-noise perturbation of the CNN prediction through a single denoising step, conditioned on the corresponding NAC PET frame via ControlNet. Finally, frame-wise predictions are averaged voxel-wise to obtain the final $\mu$-map. We further introduce the voxel-wise temporal standard deviation (TSD) to quantify inter-frame variability without requiring a reference $\mu$-map. We evaluated LUCID on an in-house dataset of 102 patients and found that it achieved the best quantitative accuracy while maintaining competitive inter-frame consistency. Ablation studies showed that refinement with a diffusion backbone pretrained on in-house CT data from the training patients was less accurate than the CNN prediction alone, highlighting the importance of large-scale CT pretraining. 
\end{abstract}


\section{Introduction}

\IEEEPARstart{P}{ositron} emission tomography (PET) is an important clinical tool for myocardial perfusion assessment, with dynamic acquisitions capturing tracer kinetics across successive time frames. Accurate quantification requires reliable attenuation correction. Clinical PET commonly uses computed tomography (CT) to obtain attenuation information, but CT acquisition adds ionizing radiation exposure. In dynamic cardiac $^{82}$Rb PET, non-simultaneous PET and CT acquisitions, together with respiratory and cardiac motion, may cause misregistration between emission images and attenuation maps ($\mu$-maps), leading to artifacts and quantitative bias~\cite{gould2007frequent,mcquaid2008sources}. CT may also be unavailable in some clinical settings. These limitations motivate CT-free attenuation correction by estimating attenuation information directly from non-attenuation-corrected (NAC) PET~\cite{lee2020review}.

Early CT-free attenuation correction methods included maximum-likelihood reconstruction of activity and attenuation (MLAA), which jointly estimates activity and attenuation from PET emission data~\cite{774167}. Time-of-flight (TOF) information improves the identifiability of this joint estimation~\cite{Defrise_2012} and has been incorporated into MLAA~\cite{6264102}. However, MLAA remains sensitive to noise and activity--attenuation crosstalk~\cite{mehranian2015joint,7493217}. Alternatively, magnetic resonance imaging (MRI) has been used to estimate $\mu$-maps through tissue classification~\cite{martinez2009tissue} or atlas-based methods~\cite{hofmann2008mri}, although conventional MRI-based approaches may inadequately represent bone and local attenuation variations~\cite{mehranian2015joint}. In recent years, deep learning has been widely applied to CT-free attenuation correction. Existing methods can be broadly categorized into two approaches: direct methods synthesize attenuation-corrected (AC) PET from NAC PET~\cite{shiri2019direct,yang2019joint,armanious2020independent,li2024learning}, while indirect methods predict $\mu$-maps or pseudo-CT from PET emission-derived inputs for subsequent physics-based attenuation correction~\cite{liu2018deep,dong2019synthetic,zhou2024pour}. Both direct and indirect approaches have employed convolutional neural networks (CNNs), generative adversarial networks (GANs), and, more recently, diffusion models~\cite{chen2025synthetic,huang2026daps,jeong2026gpdm,chen20252}. However, most studies have focused on brain or whole-body PET, while cardiac applications have primarily addressed static or gated imaging~\cite{jorgensen2026attenuation,sun2025ct,deng2026generative}. Dynamic myocardial perfusion PET remains less explored, with existing work mainly focusing on direct AC PET synthesis. Sun et al., for example, investigated frame-wise AC PET synthesis for dynamic myocardial perfusion imaging~\cite{sun2023transfer}. This direct strategy faces a particular challenge in dynamic imaging: variations in tracer distribution and noise across dynamic frames make it difficult to isolate attenuation-specific errors in synthesized AC PET. Explicit $\mu$-map prediction allows attenuation errors to be evaluated independently and frame-wise estimates to be aggregated, as the underlying attenuation structure remains approximately consistent across dynamic frames.

Although supervised CNN-based $\mu$-map synthesis provides patient-specific estimates, it relies on paired NAC PET and CT-derived $\mu$-maps, which are difficult to collect at scale due to the need for NAC reconstructions and spatially aligned CT references~\cite{deng2026generative}. Large-scale unpaired CT datasets~\cite{hamamci2026generalist,blankemeier2026merlin,huang2023inspect} offer a more accessible source of anatomical information and can be used to pretrain diffusion models to learn rich anatomical priors. The effectiveness of these priors depends on how they are incorporated into $\mu$-map synthesis, as generating from pure noise or applying excessive noise perturbations may compromise patient-specific anatomical fidelity and quantitative accuracy~\cite{webber2024diffusion}. Motivated by these considerations, we propose Large-scale Unpaired CT-Informed Diffusion (LUCID), a framework that combines anatomical priors learned from unpaired CT with patient-specific CNN predictions for CT-free $\mu$-map synthesis in dynamic cardiac $^{82}$Rb PET. Our main contributions are summarized as follows:

\begin{enumerate}
\item \textbf{Large-scale unpaired CT pretraining:} We pretrain a 3D diffusion backbone on 10,000 chest CT volumes from CT-RATE~\cite{hamamci2026generalist} to transfer anatomical priors to cardiac PET $\mu$-map synthesis. 

\item \textbf{Patient-specific single-step refinement:} We combine CNN initialization with NAC PET-conditioned ControlNet~\cite{zhang2023controlnet} refinement, using low-noise perturbation and a single deterministic denoising step to improve $\mu$-map accuracy while preserving patient-specific anatomy.

\item \textbf{Inter-frame consistency evaluation:} We aggregate frame-wise $\mu$-map predictions to obtain the final $\mu$-map and introduce the reference-free voxel-wise temporal standard deviation (TSD) to evaluate inter-frame consistency, capturing inconsistencies that may be obscured when per-frame errors are averaged over the whole image.

\end{enumerate}

Overall, LUCID outperforms all baselines in accuracy with comparable inter-frame consistency, despite the limited availability of paired PET/CT training data.
\section{Methods}

\subsection{Data Acquisition and Preprocessing}
\label{sec:data-preprocessing}

This retrospective study was approved by the Institutional Review Board of Northwestern University (Protocol No. STU00222257), with a waiver of informed consent. We identified 106 patients who underwent dynamic $^{82}$Rb myocardial perfusion PET using a Siemens Biograph Vision 600 PET/CT scanner. Three patients were excluded due to missing CT-derived $\mu$-maps and one due to missing dynamic PET data, leaving 102 patients (54 women and 48 men; mean age, $68.7 \pm 11.0$ years; median BMI, 29.5~kg/m$^2$). The dataset was split at the patient level into training, validation, and test sets of 71, 15, and 16 patients, corresponding to 137, 29, and 32 scans, respectively.

The median injected $^{82}$Rb activity was 740~MBq. Each dynamic PET scan was acquired over 430~s and reconstructed into 27 temporal frames. Following the clinical reading window at our institution, frames 19--27 (90--430~s) were selected for $\mu$-map synthesis. PET images were reconstructed using 3D time-of-flight ordered-subsets expectation maximization (TOF-OSEM) with three iterations and five subsets, without post-reconstruction filtering. The reconstructed matrix size was $220 \times 220 \times 88$, with a voxel size of $3.3 \times 3.3 \times 3.0~\mathrm{mm}^3$. NAC PET images underwent normalization, dead-time, random-coincidence, and decay corrections, without attenuation or scatter correction.

CT-derived $\mu$-maps were used as ground truth (GT) for training and evaluation. These GT $\mu$-maps were generated from low-dose CT acquired during the examination and were spatially aligned with the corresponding PET images, requiring no additional registration. Voxel-wise comparison confirmed that the GT $\mu$-maps were identical between rest and stress scans of the same patient, allowing a single GT $\mu$-map to be used for both.

For preprocessing, each NAC PET frame was converted to standardized uptake value (SUV) using patient body weight and injected activity. SUV intensities were then clipped at the 99.5th percentile and linearly scaled to $[-1,1]$. To match the HU domain used for diffusion pretraining, we fitted a piecewise linear relationship between HU and the attenuation coefficient $\mu$ at 511~keV using paired CT and GT $\mu$-map data from five training patients:

\begin{equation}
\label{eq:hu-to-mu}
\mu(h) =
\begin{cases}
C + s_1 h, & h \le 0,\\
C + s_2 h, & h > 0,
\end{cases}
\end{equation}

where $h$ denotes the HU value, $C = 0.0967~\mathrm{cm}^{-1}$, $s_1 = 9.51 \times 10^{-5}~\mathrm{cm}^{-1}/\mathrm{HU}$, and $s_2 = 3.66 \times 10^{-5}~\mathrm{cm}^{-1}/\mathrm{HU}$. GT $\mu$-maps were converted to HU using the inverse relationship, clipped to $[-1000,1500]$~HU, and linearly scaled to $[-1,1]$ as network training targets. For evaluation, network predictions were restored to the HU domain and converted back to physical attenuation coefficients using the same relationship.

\subsection{Training}

Let $\mathbf{x}^i$ denote the $i$-th dynamic NAC PET frame and
$\mathbf{y}$ the corresponding normalized GT $\mu$-map described in Sec.~\ref{sec:data-preprocessing}. During training, each $\mathbf{x}^i$ with
$i\in\mathcal{F}=\{19,\ldots,27\}$ is paired with $\mathbf{y}$ to
form an independent training sample. Each epoch processes all training
samples once in a randomly shuffled order.

\begin{figure*}[!t]
    \centering
    \includegraphics[width=0.92\textwidth]{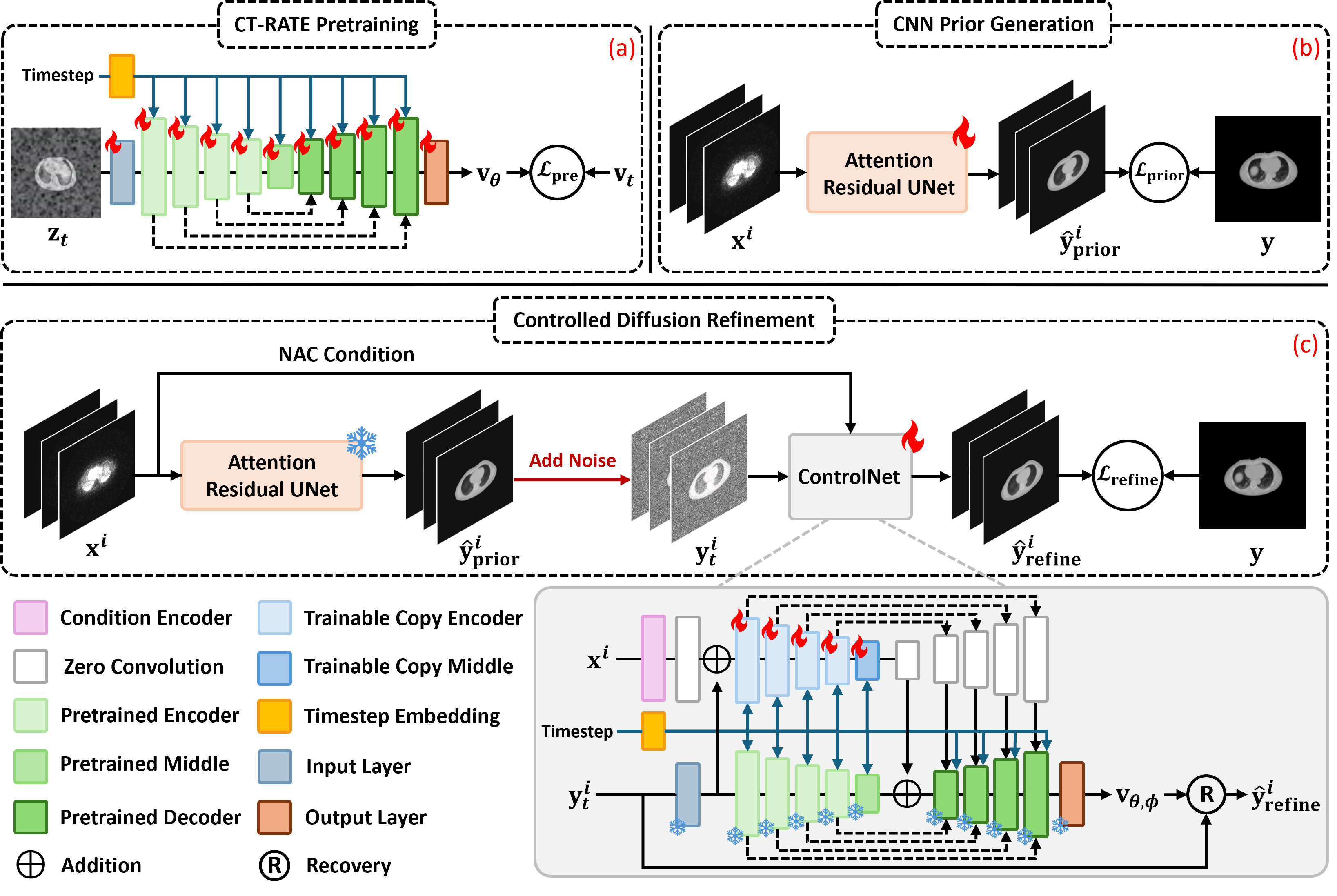}
    \caption{Overview of LUCID. (a) CT-RATE pretraining of the 3D diffusion backbone. (b) CNN prior generation. (c) Controlled diffusion refinement with the ControlNet architecture enlarged. Snowflakes and flames indicate frozen and trainable modules, respectively.}
    \label{fig:framework}
\end{figure*}

\subsubsection{Pretraining}
As shown in Fig.~\ref{fig:framework}(a), the diffusion backbone is
first pretrained on 10,000 chest CT volumes from the CT-RATE~\cite{hamamci2026generalist} dataset
to learn a 3D CT anatomical prior. The CT volumes are
resampled to the same spatial grid as the PET data and processed using
the same HU clipping and linear scaling described in Sec.~\ref{sec:data-preprocessing}. The
resulting normalized CT volume $\mathbf{z}$ is used for diffusion
pretraining.

Following the denoising diffusion probabilistic model (DDPM)~\cite{ho2020ddpm}, the forward diffusion process follows a variance schedule
$\{\beta_t\}_{t=1}^{T}$ with $T=1000$. Defining
$\alpha_t=1-\beta_t$ and
$\bar{\alpha}_t=\prod_{s=1}^{t}\alpha_s$, the noisy state
$\mathbf{z}_t$ at timestep $t$ is given by
\begin{equation}
\mathbf{z}_t
=
\sqrt{\bar{\alpha}_t}\,\mathbf{z}
+
\sqrt{1-\bar{\alpha}_t}\,\boldsymbol{\epsilon},
\qquad
\boldsymbol{\epsilon}\sim
\mathcal{N}(\mathbf{0},\mathbf{I}).
\label{eq:pre_forward}
\end{equation}

The network adopts the $v$-parameterization~\cite{salimans2022progressive}, with the velocity target
$\mathbf{v}_t$ defined as
\begin{equation}
\mathbf{v}_t
=
\sqrt{\bar{\alpha}_t}\,\boldsymbol{\epsilon}
-
\sqrt{1-\bar{\alpha}_t}\,\mathbf{z}.
\label{eq:v_target}
\end{equation}

Under the $v$-parameterization,
$\mathbf{z}$ can be recovered from $\mathbf{z}_t$ and $\mathbf{v}_t$
as
\begin{equation}
\mathbf{z}
=
\sqrt{\bar{\alpha}_t}\,\mathbf{z}_t
-
\sqrt{1-\bar{\alpha}_t}\,\mathbf{v}_t.
\label{eq:v_recovery}
\end{equation}

The denoising network $\mathbf{v}_{\theta}$ is implemented as a
3D UNet with timestep embeddings, where $\theta$ denotes the
backbone parameters. It is optimized using
\begin{equation}
\mathcal{L}_{\mathrm{pre}}
=
\mathbb{E}
\left[
\left\|
\mathbf{v}_{\theta}(\mathbf{z}_t,t)
-
\mathbf{v}_t
\right\|_2^2
\right].
\label{eq:pre_loss}
\end{equation}

After pretraining, $\theta$ remains frozen throughout the subsequent
training stages.

\subsubsection{CNN Prior Generation}
As shown in Fig.~\ref{fig:framework}(b), a 3D Attention Residual UNet~\cite{cciccek20163d,he2016deep}
$f_{\mathrm{prior}}$ predicts an initial $\mu$-map from each
$\mathbf{x}^i$. The network uses a five-level encoder--decoder
architecture, with four-head self-attention~\cite{vaswani2017attention} applied only at the fourth
level and the bottleneck. The initial frame-wise prediction
$\hat{\mathbf{y}}_{\mathrm{prior}}^i$ is defined as
\begin{equation}
\hat{\mathbf{y}}_{\mathrm{prior}}^i
=
f_{\mathrm{prior}}(\mathbf{x}^i),
\label{eq:prior_prediction}
\end{equation}
and is optimized using
\begin{equation}
\mathcal{L}_{\mathrm{prior}}
=
\mathbb{E}
\left[
\left\|
\hat{\mathbf{y}}_{\mathrm{prior}}^i
-
\mathbf{y}
\right\|_1
\right].
\label{eq:prior_loss}
\end{equation}

All temporal frames from the same scan share the same $\mathbf{y}$.
After training, $f_{\mathrm{prior}}$ is frozen, and its output is used
to initialize the subsequent diffusion refinement stage.

\subsubsection{Controlled Diffusion Refinement}
Fig.~\ref{fig:framework}(c) illustrates the controlled diffusion
refinement stage. Both $\theta$ and $f_{\mathrm{prior}}$ remain
frozen, while only the ControlNet~\cite{zhang2023controlnet} parameters $\phi$ are optimized.
ControlNet consists of trainable copies of the backbone encoder and
middle block and is conditioned on $\mathbf{x}^i$. The conditioning
input is first processed by a condition encoder and a zero convolution,
and the resulting features are added to the output of the trainable
input layer. Features from each encoder level and the middle block are
then passed through separate zero convolutions and injected into the
corresponding features of the frozen backbone. This design introduces
NAC PET conditioning while preserving the anatomical prior learned
during CT pretraining.

Rather than starting from pure noise, the refinement process is
initialized from $\hat{\mathbf{y}}_{\mathrm{prior}}^i$. During
training, the timestep is sampled uniformly from the discrete
low-noise range
\begin{equation}
t
\sim
\mathcal{U}
\left\{
t_{\min},\ldots,t_{\max}-1
\right\},
\label{eq:t_sampling}
\end{equation}
where $t_{\min}=20$ and $t_{\max}=100$. The corresponding noisy state
$\mathbf{y}_t^i$ is constructed as
\begin{equation}
\mathbf{y}_t^i
=
\sqrt{\bar{\alpha}_t}\,
\hat{\mathbf{y}}_{\mathrm{prior}}^i
+
\sqrt{1-\bar{\alpha}_t}\,
\boldsymbol{\epsilon}.
\label{eq:refine_forward}
\end{equation}

The lower bound excludes very small $t$, where $\mathbf{y}_t^i$ is
nearly identical to $\hat{\mathbf{y}}_{\mathrm{prior}}^i$ and the
contribution of the network prediction in Eq.~\eqref{eq:refine_recovery}
is strongly attenuated by the small factor
$\sqrt{1-\bar{\alpha}_t}$, limiting the correction of residual errors
in the initialization.

The conditional denoising network
$\mathbf{v}_{\theta,\phi}$ combines the frozen backbone parameters
$\theta$ with the trainable ControlNet parameters $\phi$. Following
the recovery relation in Eq.~\eqref{eq:v_recovery}, the refined
$\mu$-map $\hat{\mathbf{y}}_{\mathrm{refine}}^i$, corresponding to
the recovery operator \textcircled{\scriptsize R} in Fig.~\ref{fig:framework}(c), is given by
\begin{equation}
\hat{\mathbf{y}}_{\mathrm{refine}}^i
=
\sqrt{\bar{\alpha}_t}\,
\mathbf{y}_t^i
-
\sqrt{1-\bar{\alpha}_t}\,
\mathbf{v}_{\theta,\phi}
\left(
\mathbf{y}_t^i,t,\mathbf{x}^i
\right).
\label{eq:refine_recovery}
\end{equation}

Because $\mathbf{y}_t^i$ is constructed from
$\hat{\mathbf{y}}_{\mathrm{prior}}^i$ rather than $\mathbf{y}$,
the explicit $v$ target defined from $\mathbf{y}$ and
$\boldsymbol{\epsilon}$ in standard diffusion training is no longer
strictly matched to $\mathbf{y}_t^i$. We therefore retain the
$v$-parameterized output while directly supervising
$\hat{\mathbf{y}}_{\mathrm{refine}}^i$ in the image domain:
\begin{equation}
\mathcal{L}_{\mathrm{refine}}
=
\mathbb{E}
\left[
\left\|
\hat{\mathbf{y}}_{\mathrm{refine}}^i
-
\mathbf{y}
\right\|_1
\right].
\label{eq:refine_loss}
\end{equation}

Only $\phi$ is updated by this loss. Throughout training, the diffusion
starting state is always constructed from the output of the frozen
$f_{\mathrm{prior}}$, maintaining consistency between refinement
training and inference.

\subsection{Inference}

During inference, frame-wise refinement follows the same pipeline
shown in Fig.~\ref{fig:framework}(c), with the starting timestep fixed
at $t_s=25$. For each $\mathbf{x}^i$, $f_{\mathrm{prior}}$ first
produces $\hat{\mathbf{y}}_{\mathrm{prior}}^i$, from which the initial
noisy state $\mathbf{y}_{t_s}^i$ is constructed as
\begin{equation}
\mathbf{y}_{t_s}^i
=
\sqrt{\bar{\alpha}_{t_s}}\,
\hat{\mathbf{y}}_{\mathrm{prior}}^i
+
\sqrt{1-\bar{\alpha}_{t_s}}\,
\boldsymbol{\epsilon}.
\label{eq:inference_initialization}
\end{equation}

Starting from $\mathbf{y}_{t_s}^i$, we apply Denoising Diffusion
Implicit Models (DDIM)~\cite{song2021denoising} sampling with
$\eta=0$ over $K$ uniformly spaced steps down to $t=0$, and obtain the
refined prediction $\hat{\mathbf{y}}_{\mathrm{refine}}^i$ from
Eq.~\eqref{eq:refine_recovery}. With $\eta=0$, the reverse process adds
no noise, so the only randomness comes from the noise added to the CNN
prior. We use $K=1$ by default to keep distortion low, as a single step
has been observed to give the lowest distortion, while more steps
improve perceptual quality at the cost of
fidelity~\cite{delbracio2023inversion}.

Because all temporal frames within the same scan share the same
underlying attenuation structure,
$\hat{\mathbf{y}}_{\mathrm{refine}}^i$ for
$i\in\mathcal{F}$ are averaged voxel-wise to obtain the final
scan-level $\mu$-map $\hat{\mathbf{y}}$:
\begin{equation}
\hat{\mathbf{y}}
=
\frac{1}{|\mathcal{F}|}
\sum_{i\in\mathcal{F}}
\hat{\mathbf{y}}_{\mathrm{refine}}^i.
\label{eq:temporal_aggregation}
\end{equation}

\subsection{Model Implementation}

The proposed method was implemented in PyTorch and trained on a
single NVIDIA A100 GPU with 80 GB memory. All training stages used
the AdamW optimizer, mixed-precision training, and gradient clipping,
with early stopping based on validation performance. Both diffusion
backbone pretraining and ControlNet training used a learning rate of
$10^{-4}$ and a batch size of 2, with an exponential moving average
(EMA) decay of 0.9999 applied to the model weights. The Attention Residual
UNet was trained with a learning rate of $2\times10^{-4}$ and a
batch size of 4. Training was limited to a maximum of 200 epochs.

\subsection{Evaluation Metrics and Baselines}
\label{sec:evaluation}

For quantitative evaluation, mean absolute error (MAE), peak signal-to-noise ratio (PSNR), and structural similarity index (SSIM) were computed over all voxels after converting both predictions and GT $\mu$-maps back to the physical attenuation domain. MAE and PSNR assess the recovery of attenuation values, while SSIM evaluates structural similarity.

To evaluate the inter-frame consistency of dynamic $\mu$-map synthesis, we propose voxel-wise temporal standard deviation (TSD). Since dynamic frames share approximately the same attenuation structure, variations in frame-wise predictions reflect the model's sensitivity to frame-dependent input characteristics, including tracer distribution and noise levels. TSD measures this variation by computing the standard deviation of predictions at each voxel and averaging over all voxels:

\begin{equation}
\mathrm{TSD}
=
\frac{1}{|\Omega|}
\sum_{j\in\Omega}
\sqrt{
\frac{1}{|\mathcal{F}|}
\sum_{i\in\mathcal{F}}
\left(
\hat{\mu}^i_j-\bar{\mu}_j
\right)^2
},
\label{eq:tsd}
\end{equation}

where $\mathcal{F}$ denotes the selected dynamic frames (19--27), $\hat{\mu}^i_j$ is the attenuation coefficient predicted from frame $i$ at voxel $j$, $\bar{\mu}_j$ is the mean prediction across frames at voxel $j$, and $\Omega$ denotes all evaluated voxels. TSD shares the same units as MAE, with lower values indicating greater inter-frame consistency. Statistical significance was assessed using a two-sided paired Wilcoxon signed-rank test on per-scan results, with $p<0.05$ considered significant.

We compared LUCID with four baseline methods: 3D UNet~\cite{cciccek20163d}, Attn Pix2Pix~\cite{jorgensen2026attenuation}, BBDM~\cite{li2023bbdm}, and DDPM~\cite{ho2020ddpm}. All methods used the same patient-level data split, preprocessing pipeline, and frames 19--27. Frame-wise predictions were averaged voxel-wise to obtain the final $\mu$-map.

\section{Results}

\subsection{Comparison with Existing Methods}
\label{sec:baselines}

\begin{figure*}[!t]
\centering
\includegraphics[width=0.92\textwidth]{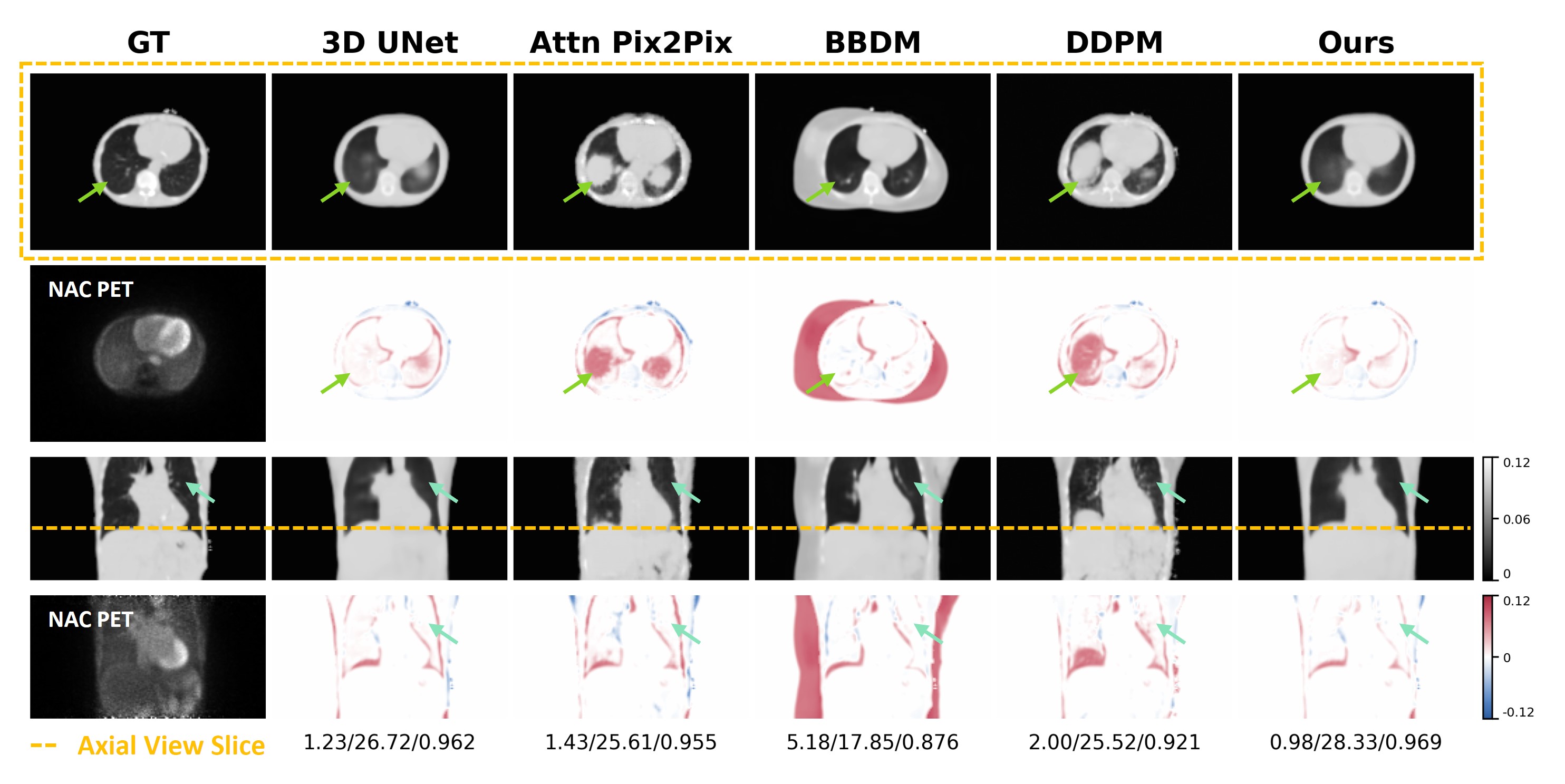}
\caption{Qualitative comparison of synthesized $\mu$-maps for a representative test scan. Scan-level MAE, PSNR, and SSIM are reported below each column.}
\label{fig:main_comparison}
\end{figure*}

Fig.~\ref{fig:main_comparison} presents the qualitative results, showing synthesized $\mu$-maps and corresponding error maps for a representative test scan. The green arrows highlight a region near the hepatic dome that corresponds to lung tissue in the GT $\mu$-map. Most methods incorrectly generated additional liver-like soft tissue in this region, producing clear positive errors. This overestimation was most pronounced for Attn Pix2Pix~\cite{jorgensen2026attenuation} and DDPM~\cite{ho2020ddpm}, whereas LUCID showed the smallest error.

Differences were also evident along the body contour. Attn Pix2Pix produced noticeable jagged artifacts at the body boundary, while BBDM~\cite{li2023bbdm} substantially overestimated the body contour, resulting in continuous bands of positive error in both axial and coronal views and an MAE of $5.18\times10^{-3}\,\mathrm{cm}^{-1}$ for this scan. Attn Pix2Pix, BBDM, and DDPM also generated rich local textures within the lungs, as indicated by the cyan arrows. Although visually plausible, these details did not correspond to the anatomy in the GT $\mu$-map and therefore introduced additional errors.

Among the comparison methods, 3D UNet~\cite{cciccek20163d} most closely approached LUCID, recovering the overall anatomy with substantially less excess soft tissue near the hepatic dome and achieving an MAE of $1.23\times10^{-3}\,\mathrm{cm}^{-1}$. However, its lung structures remained relatively blurred, with more positive errors along the lung and body boundaries, indicating less accurate boundary recovery. In contrast, LUCID better preserved both the lung and body contours relative to the GT $\mu$-map, with residual errors largely confined to narrow structural boundaries. It achieved an MAE of $0.98\times10^{-3}\,\mathrm{cm}^{-1}$ for this scan, lower than all comparison methods.

\begin{table*}[!t]
\centering
\caption{Quantitative comparison with baseline methods on the test set. Results are reported as mean $\pm$ standard deviation across 16 rest and 16 stress scans. MAE and TSD are in $10^{-3}$\,cm$^{-1}$, and PSNR is in dB. Best and second-best results are shown in \fst{red} and \snd{blue}, respectively. $\ast$ and $\dagger$ indicate significantly worse and better performance than LUCID, respectively.}
\label{tab:main}
\setlength{\tabcolsep}{4pt}
\renewcommand{\arraystretch}{1.15}
\resizebox{\textwidth}{!}{%
\begin{tabular}{l cccc cccc}
\toprule
\multirow{2}{*}[-0.5ex]{Method} & \multicolumn{4}{c}{Rest} & \multicolumn{4}{c}{Stress} \\
\cmidrule(lr){2-5}\cmidrule(lr){6-9}
 & MAE $\downarrow$ & PSNR $\uparrow$ & SSIM $\uparrow$ & TSD $\downarrow$
 & MAE $\downarrow$ & PSNR $\uparrow$ & SSIM $\uparrow$ & TSD $\downarrow$ \\
\midrule
3D UNet~\cite{cciccek20163d}
 & 1.52$\pm$0.85\rlap{$^{*}$} & 27.58$\pm$2.89\rlap{$^{*}$} & \snd{0.938$\pm$0.039}\rlap{$^{*}$} & \snd{0.500$\pm$0.280}\rlap{$^{\dagger}$}
 & 1.63$\pm$0.88\rlap{$^{*}$} & 26.85$\pm$2.74\rlap{$^{*}$} & \snd{0.932$\pm$0.041}\rlap{$^{*}$} & \fst{0.559$\pm$0.298}\rlap{$^{\dagger}$} \\
Attn Pix2Pix~\cite{jorgensen2026attenuation}
 & \snd{1.49$\pm$0.70}\rlap{$^{*}$} & \snd{27.71$\pm$2.48}\rlap{$^{*}$} & 0.935$\pm$0.037\rlap{$^{*}$} & 0.617$\pm$0.302\rlap{$^{*}$}
 & \snd{1.62$\pm$0.73}\rlap{$^{*}$} & \snd{26.94$\pm$2.36}\rlap{$^{*}$} & 0.929$\pm$0.039\rlap{$^{*}$} & 0.691$\pm$0.323\rlap{$^{*}$} \\
BBDM~\cite{li2023bbdm}
 & 2.25$\pm$1.25\rlap{$^{*}$} & 25.50$\pm$3.70\rlap{$^{*}$} & 0.915$\pm$0.042\rlap{$^{*}$} & 0.682$\pm$0.480
 & 2.42$\pm$1.24\rlap{$^{*}$} & 24.70$\pm$3.27\rlap{$^{*}$} & 0.907$\pm$0.043\rlap{$^{*}$} & 0.800$\pm$0.589 \\
DDPM~\cite{ho2020ddpm}
 & 2.34$\pm$0.91\rlap{$^{*}$} & 26.36$\pm$2.71\rlap{$^{*}$} & 0.886$\pm$0.045\rlap{$^{*}$} & \fst{0.490$\pm$0.257}
 & 2.44$\pm$0.93\rlap{$^{*}$} & 25.79$\pm$2.46\rlap{$^{*}$} & 0.883$\pm$0.046\rlap{$^{*}$} & \snd{0.574$\pm$0.265} \\
\midrule
Ours
 & \fst{1.37$\pm$0.77} & \fst{28.44$\pm$2.87} & \fst{0.944$\pm$0.036} & 0.515$\pm$0.278
 & \fst{1.49$\pm$0.78} & \fst{27.63$\pm$2.65} & \fst{0.939$\pm$0.037} & 0.578$\pm$0.283 \\
\bottomrule
\end{tabular}}
\end{table*}

Table~\ref{tab:main} summarizes the quantitative results on the test set. LUCID achieved the lowest MAE and the highest PSNR and SSIM for both rest and stress scans, with significant improvements over all comparison methods across all three metrics (all $p<0.01$).

Compared with 3D UNet and Attn Pix2Pix, LUCID showed consistent gains in both phases. For rest scans, it achieved an MAE of $1.37\times10^{-3}\,\mathrm{cm}^{-1}$, representing reductions of approximately 8\% and 10\% relative to Attn Pix2Pix and 3D UNet, respectively, while improving PSNR by 0.73 and 0.86~dB. For stress scans, the MAE was $1.49\times10^{-3}\,\mathrm{cm}^{-1}$, corresponding to reductions of approximately 8\% and 9\%, with PSNR gains of 0.69 and 0.78~dB. 3D UNet was the second-best method in SSIM, reaching 0.938 and 0.932 for rest and stress scans, respectively, compared with 0.944 and 0.939 for LUCID.

The improvement was more pronounced over the diffusion baselines. DDPM and BBDM both yielded MAEs above $2.2\times10^{-3}\,\mathrm{cm}^{-1}$, whereas LUCID reduced MAE by approximately 40\% and improved PSNR by about 2--3~dB. Their SSIM values reached at most 0.915, compared with 0.944 and 0.939 for LUCID, with DDPM showing the lowest SSIM in both phases. BBDM also exhibited substantially larger standard deviations in MAE and PSNR than LUCID. Overall, 3D UNet and Attn Pix2Pix outperformed the diffusion baselines, while LUCID further improved synthesis accuracy in both phases. Although MAE increased from rest to stress for all methods, the relative advantage of LUCID was maintained.

\subsection{Frame-wise Consistency Analysis}
\label{sec:consistenc}

\begin{figure*}[!t]
\centering
\includegraphics[width=\textwidth]{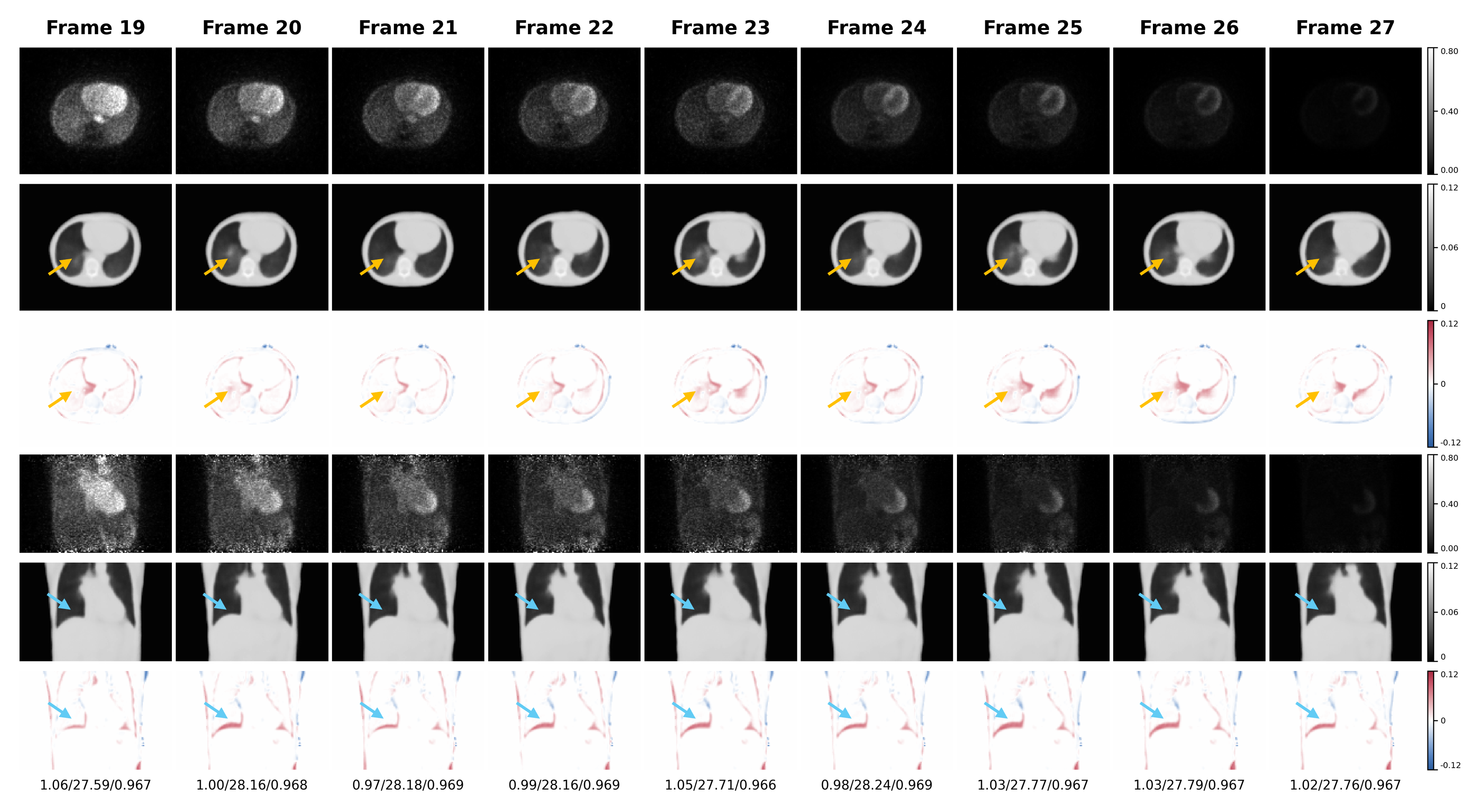}
\caption{Frame-wise $\mu$-map synthesis for frames 19--27 of a representative test scan. Axial and coronal views show the NAC PET input, predicted $\mu$-map, and error map. Frame-wise MAE, PSNR, and SSIM are reported below each column.}
\label{fig:dynamic}
\end{figure*}

For qualitative evaluation, Fig.~\ref{fig:dynamic} shows the frame-wise $\mu$-maps synthesized by LUCID from frames 19--27 of the same rest scan shown in Fig.~\ref{fig:main_comparison}, with axial and coronal views. The NAC PET inputs change markedly across frames: early frames show high blood-pool activity, followed by progressively lower counts and increasing localization of tracer uptake to the left ventricular myocardium. Despite these temporal changes, the synthesized $\mu$-maps preserve consistent body contours, lung boundaries, and spinal anatomy across frames.

The main frame-to-frame variation occurs at the right lung base--hepatic dome interface, highlighted by the orange arrows in the axial views and the blue arrows in the coronal views. The extent of additional liver-like soft tissue in this region varies across frames and becomes more apparent in later frames. Elsewhere, the error pattern remains largely stable, with positive errors near the lung boundaries and negative errors along the body contour. Overall, LUCID maintains stable attenuation anatomy despite substantial changes in tracer distribution and count level, with residual temporal variation largely confined to local tissue boundaries.

For quantitative evaluation of frame-to-frame consistency, all TSD values are reported in units of $10^{-3}\,\mathrm{cm}^{-1}$. LUCID achieved TSD values of 0.515 and 0.578 for the rest and stress scans, respectively, significantly lower than those of Attn Pix2Pix at 0.617 and 0.691, with both $p<0.001$. 3D UNet achieved significantly lower TSD values than LUCID for both rest ($p=0.044$) and stress ($p=0.013$) scans, while BBDM and DDPM showed no significant differences. However, all three methods exhibited significantly higher MAE, suggesting that temporally consistent predictions may still contain persistent errors. Overall, LUCID improved $\mu$-map synthesis accuracy while maintaining competitive frame-to-frame consistency.

\subsection{Ablation Studies}
\label{sec:ablation}

\subsubsection{Component ablation} 
\begin{figure}[t]
    \centering
    \includegraphics[width=\linewidth]{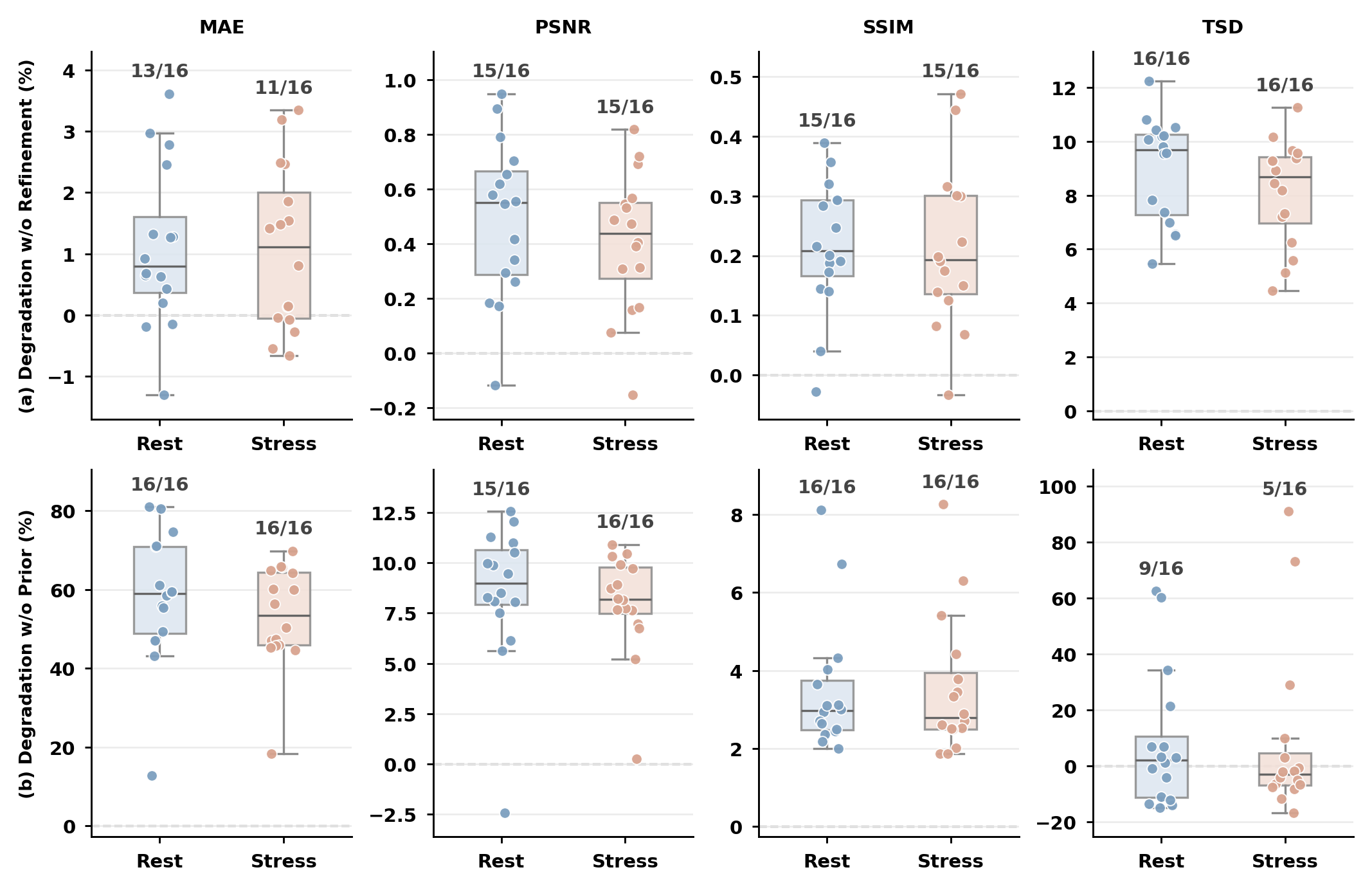}
    \caption{Per-scan performance degradation in the component ablation under the Frame-wise strategy.}
    \label{fig:ablation_delta}
\end{figure}

To evaluate the roles of CNN initialization and diffusion refinement, we compared LUCID with two variants. The w/o Refinement variant directly uses the initial $\mu$-map predicted by the Attention Residual UNet~\cite{cciccek20163d,he2016deep,vaswani2017attention}, without diffusion refinement. The w/o Prior variant generates $\mu$-maps from pure noise without CNN initialization, with ControlNet trained across all diffusion timesteps and 1000 sampling steps. Quantitative results are presented in Table~\ref{tab:ablation}, while Fig.~\ref{fig:ablation_delta} shows the per-scan performance degradation relative to LUCID. Degradation is calculated as the relative increase in MAE or the relative decrease in PSNR and SSIM, with positive values indicating worse performance.

\providecommand{\fst}[1]{\textcolor{red}{#1}}
\providecommand{\snd}[1]{\textcolor{blue}{#1}}

\begin{table*}[!t]
\centering
\caption{Ablation results on the test set. Results are reported as mean $\pm$ standard deviation across 16 rest and 16 stress scans. MAE and TSD are in $10^{-3}$\,cm$^{-1}$, and PSNR is in dB. Best and second-best results are shown in \fst{red} and \snd{blue}, respectively. $^{*}$ indicates significantly worse performance than Frame-wise Ours.}
\label{tab:ablation}

\setlength{\tabcolsep}{4pt}
\renewcommand{\arraystretch}{1.15}
\resizebox{\textwidth}{!}{%
\begin{tabular}{ll cccc cccc}
\toprule
\multirow{2}{*}[-0.5ex]{Frame usage} & \multirow{2}{*}[-0.5ex]{Method} & \multicolumn{4}{c}{Rest} & \multicolumn{4}{c}{Stress} \\
\cmidrule(lr){3-6}\cmidrule(lr){7-10}
 & & MAE $\downarrow$ & PSNR $\uparrow$ & SSIM $\uparrow$ & TSD $\downarrow$
 & MAE $\downarrow$ & PSNR $\uparrow$ & SSIM $\uparrow$ & TSD $\downarrow$ \\
\midrule

\multirow{3}{*}{Single frame}
 & w/o Refinement
 & 1.49$\pm$0.82\rlap{$^{*}$}
 & 27.41$\pm$2.75\rlap{$^{*}$}
 & 0.935$\pm$0.041\rlap{$^{*}$}
 & --
 & 1.61$\pm$0.78\rlap{$^{*}$}
 & 26.60$\pm$2.53\rlap{$^{*}$}
 & 0.930$\pm$0.039\rlap{$^{*}$}
 & -- \\

 & w/o Prior
 & 2.19$\pm$1.09\rlap{$^{*}$}
 & 25.33$\pm$2.56\rlap{$^{*}$}
 & 0.907$\pm$0.050\rlap{$^{*}$}
 & --
 & 2.29$\pm$1.13\rlap{$^{*}$}
 & 24.89$\pm$2.43\rlap{$^{*}$}
 & 0.903$\pm$0.049\rlap{$^{*}$}
 & -- \\

 & Ours
 & 1.46$\pm$0.81\rlap{$^{*}$}
 & 27.64$\pm$2.82\rlap{$^{*}$}
 & 0.938$\pm$0.040\rlap{$^{*}$}
 & --
 & 1.58$\pm$0.77\rlap{$^{*}$}
 & 26.82$\pm$2.58\rlap{$^{*}$}
 & 0.933$\pm$0.038\rlap{$^{*}$}
 & -- \\

\midrule

\multirow{3}{*}{Input averaging}
 & w/o Refinement
 & \snd{1.38$\pm$0.82}
 & 28.13$\pm$2.94\rlap{$^{*}$}
 & \snd{0.942$\pm$0.038}\rlap{$^{*}$}
 & --
 & 1.51$\pm$0.84
 & 27.29$\pm$2.80\rlap{$^{*}$}
 & 0.936$\pm$0.039\rlap{$^{*}$}
 & -- \\

 & w/o Prior
 & 2.04$\pm$0.98\rlap{$^{*}$}
 & 25.92$\pm$2.51\rlap{$^{*}$}
 & 0.914$\pm$0.044\rlap{$^{*}$}
 & --
 & 2.17$\pm$1.02\rlap{$^{*}$}
 & 25.36$\pm$2.37\rlap{$^{*}$}
 & 0.908$\pm$0.046\rlap{$^{*}$}
 & -- \\

 & Ours
 & \fst{1.37$\pm$0.82}
 & 28.29$\pm$3.02
 & \fst{0.944$\pm$0.037}
 & --
 & \fst{1.49$\pm$0.84}
 & 27.43$\pm$2.85\rlap{$^{*}$}
 & \snd{0.938$\pm$0.038}
 & -- \\

\midrule

\multirow{3}{*}{Frame-wise}
 & w/o Refinement
 & 1.39$\pm$0.77\rlap{$^{*}$}
 & \snd{28.30$\pm$2.81}\rlap{$^{*}$}
 & \snd{0.942$\pm$0.037}\rlap{$^{*}$}
 & \snd{0.558$\pm$0.290}\rlap{$^{*}$}
 & \snd{1.50$\pm$0.78}\rlap{$^{*}$}
 & \snd{27.52$\pm$2.62}\rlap{$^{*}$}
 & 0.937$\pm$0.038\rlap{$^{*}$}
 & \snd{0.621$\pm$0.294}\rlap{$^{*}$} \\

 & w/o Prior
 & 2.10$\pm$1.04\rlap{$^{*}$}
 & 25.98$\pm$2.42\rlap{$^{*}$}
 & 0.911$\pm$0.047\rlap{$^{*}$}
 & 0.585$\pm$0.399
 & 2.25$\pm$1.12\rlap{$^{*}$}
 & 25.42$\pm$2.41\rlap{$^{*}$}
 & 0.906$\pm$0.048\rlap{$^{*}$}
 & 0.659$\pm$0.464 \\

 & Ours
 & \fst{1.37$\pm$0.77}
 & \fst{28.44$\pm$2.87}
 & \fst{0.944$\pm$0.036}
 & \fst{0.515$\pm$0.278}
 & \fst{1.49$\pm$0.78}
 & \fst{27.63$\pm$2.65}
 & \fst{0.939$\pm$0.037}
 & \fst{0.578$\pm$0.283} \\

\bottomrule
\end{tabular}}
\end{table*}

Without diffusion refinement, MAE increased slightly from 1.37 to 1.39 $\times 10^{-3}$~cm$^{-1}$ for rest scans and from 1.49 to 1.50 $\times 10^{-3}$~cm$^{-1}$ for stress scans, with statistically significant differences. As shown in Fig.~\ref{fig:ablation_delta}(a), MAE increased in 13/16 rest and 11/16 stress scans, while PSNR and SSIM decreased in 15/16 scans of each series. The improvement in inter-frame consistency was more pronounced. LUCID achieved TSD values of 0.515 and 0.578, compared with 0.558 and 0.621 $\times 10^{-3}$~cm$^{-1}$ without refinement. TSD improved in all 32 scans ($p=3.05\times10^{-5}$ for both series).

Without CNN initialization, MAE increased to 2.10 and 2.25 $\times 10^{-3}$~cm$^{-1}$ for rest and stress scans, respectively, while PSNR and SSIM decreased significantly. As shown in Fig.~\ref{fig:ablation_delta}(b), MAE increased in all 32 scans, with degradation approaching 80\% in some cases. Mean TSD was higher, but the difference was not statistically significant. These results indicate that CNN initialization provides the main accuracy benefit, while diffusion refinement improves accuracy and inter-frame consistency.

\subsubsection{Frame Usage Strategy} 
\begin{figure}[t]
    \centering
    \includegraphics[width=\linewidth]{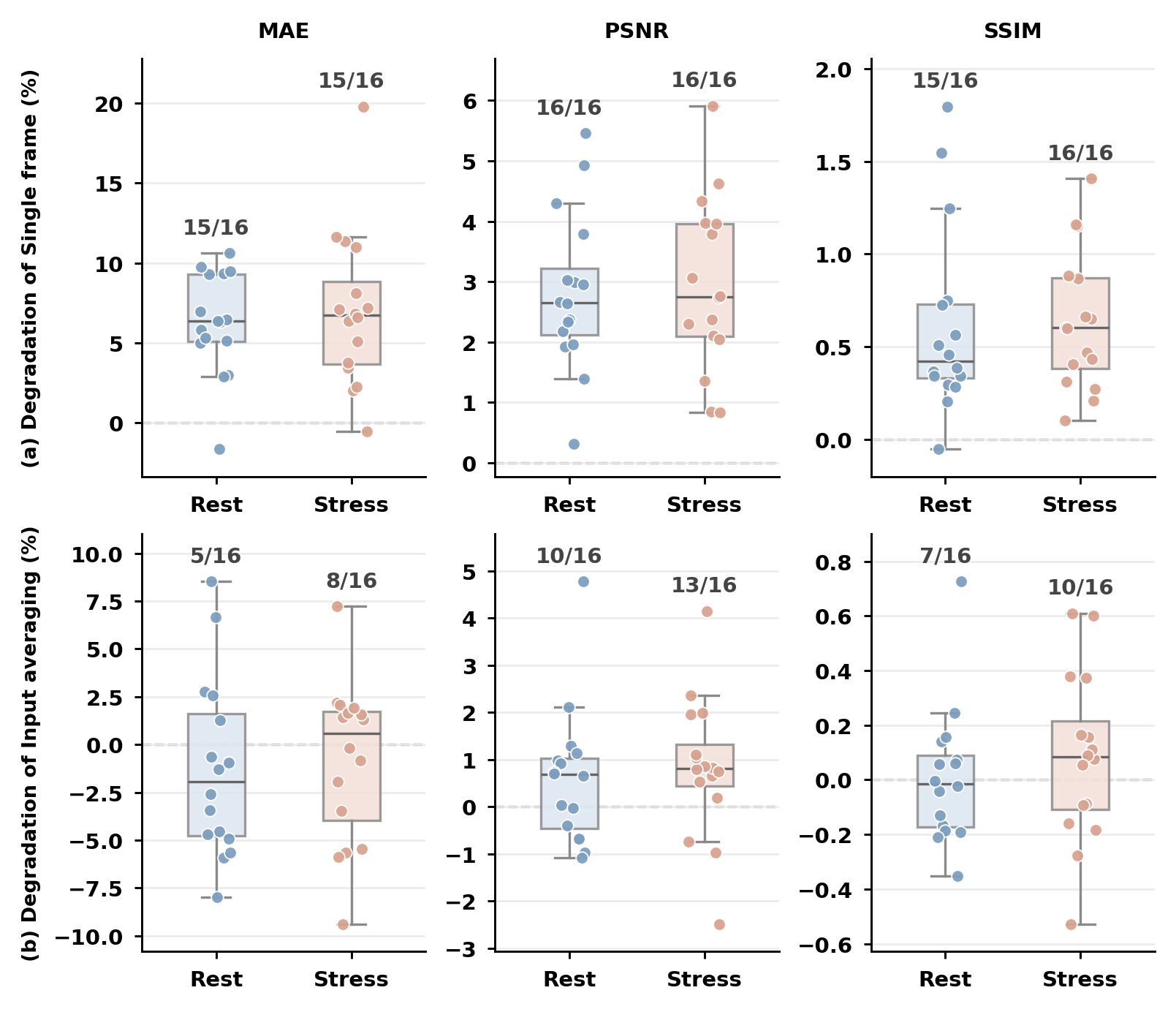}
    \caption{Per-scan performance degradation of LUCID under different frame-usage strategies relative to the Frame-wise strategy.}
    \label{fig:ablation_frame}
\end{figure}

We compared three strategies: Single frame, using only frame 23; Input averaging, averaging NAC PET frames 19--27 before synthesis; and Frame-wise, processing each frame independently and averaging the predicted $\mu$-maps. Fig.~\ref{fig:ablation_frame} shows the per-scan performance degradation of the first two strategies relative to Frame-wise, using the same percentage calculation as Fig.~\ref{fig:ablation_delta}.

Using the Single frame strategy, MAE increased from 1.37 to 1.46 $\times 10^{-3}$~cm$^{-1}$ for rest scans and from 1.49 to 1.58 $\times 10^{-3}$~cm$^{-1}$ for stress scan, with significant reductions in PSNR and SSIM. As shown in Fig.~\ref{fig:ablation_frame}, PSNR decreased in all 32 scans, while MAE degradation approached 20\% in one stress scan. The Input averaging and Frame-wise strategies achieved similar performance, with identical mean MAE values and no significant differences in MAE or SSIM. Frame-wise achieved slightly higher PSNR for stress scans ($p=0.044$). These results indicate that the improvement mainly comes from using multiple dynamic frames, with little difference between the two averaging strategies. We selected the Frame-wise strategy because it preserves individual frame predictions, enabling reference-free evaluation of inter-frame consistency using TSD.

\subsubsection{Pretraining Data} To evaluate the effect of large-scale CT pretraining data, we pretrained the diffusion backbone using in-house chest CT scans from the training patients, with the same settings as CT-RATE pretraining. Separate ControlNets were trained for pure-noise and CNN-initialized generation. Table~\ref{tab:pretrain} presents the results, including the Frame-wise CNN prior as a reference.

\providecommand{\fst}[1]{\textcolor{red}{#1}}
\providecommand{\snd}[1]{\textcolor{blue}{#1}}

\begin{table*}[!t]
\centering
\caption{Effect of the pretraining data for the diffusion backbone on
the test set. Results are reported as mean $\pm$ standard deviation
across 16 rest and 16 stress scans. MAE and TSD are in
$10^{-3}$\,cm$^{-1}$, and PSNR is in dB. Best and second-best results
are shown in \fst{red} and \snd{blue}, respectively. $^{*}$ indicates
significantly worse performance than CT-RATE counterpart with the
same initialization.}
\label{tab:pretrain}

\setlength{\tabcolsep}{4pt}
\renewcommand{\arraystretch}{1.15}
\resizebox{\textwidth}{!}{%
\begin{tabular}{lc cccc cccc}
\toprule
\multirow{2}{*}[-0.5ex]{Method} & \multirow{2}{*}[-0.5ex]{Pretraining} & \multicolumn{4}{c}{Rest} & \multicolumn{4}{c}{Stress} \\
\cmidrule(lr){3-6}\cmidrule(lr){7-10}
 & & MAE $\downarrow$ & PSNR $\uparrow$ & SSIM $\uparrow$ & TSD $\downarrow$
 & MAE $\downarrow$ & PSNR $\uparrow$ & SSIM $\uparrow$ & TSD $\downarrow$ \\
\midrule

CNN prior & --
 & \snd{1.39$\pm$0.77}
 & 28.30$\pm$2.81
 & \snd{0.942$\pm$0.037}
 & 0.558$\pm$0.290
 & \snd{1.50$\pm$0.78}
 & 27.52$\pm$2.62
 & \snd{0.937$\pm$0.038}
 & 0.621$\pm$0.294 \\

\midrule

\multirow{2}{*}{Pure-noise} & In-house
 & 2.35$\pm$0.87\rlap{$^{*}$}
 & 26.13$\pm$2.29
 & 0.889$\pm$0.042\rlap{$^{*}$}
 & \snd{0.530$\pm$0.264}
 & 2.46$\pm$0.92\rlap{$^{*}$}
 & 25.60$\pm$2.20
 & 0.885$\pm$0.044\rlap{$^{*}$}
 & 0.613$\pm$0.320 \\

 & CT-RATE
 & 2.10$\pm$1.04
 & 25.98$\pm$2.42
 & 0.911$\pm$0.047
 & 0.585$\pm$0.399
 & 2.25$\pm$1.12
 & 25.42$\pm$2.41
 & 0.906$\pm$0.048
 & 0.659$\pm$0.464 \\

\midrule

\multirow{2}{*}{Ours} & In-house
 & 1.46$\pm$0.76\rlap{$^{*}$}
 & \snd{28.38$\pm$2.85}\rlap{$^{*}$}
 & 0.941$\pm$0.037\rlap{$^{*}$}
 & \snd{0.530$\pm$0.281}\rlap{$^{*}$}
 & 1.58$\pm$0.76\rlap{$^{*}$}
 & \snd{27.55$\pm$2.62}\rlap{$^{*}$}
 & 0.936$\pm$0.037\rlap{$^{*}$}
 & \snd{0.592$\pm$0.285}\rlap{$^{*}$} \\

 & CT-RATE
 & \fst{1.37$\pm$0.77}
 & \fst{28.44$\pm$2.87}
 & \fst{0.944$\pm$0.036}
 & \fst{0.515$\pm$0.278}
 & \fst{1.49$\pm$0.78}
 & \fst{27.63$\pm$2.65}
 & \fst{0.939$\pm$0.037}
 & \fst{0.578$\pm$0.283} \\

\bottomrule
\end{tabular}}
\end{table*}

Compared with CT-RATE pretraining, in-house pretraining significantly worsened all four metrics for LUCID. For pure-noise generation, MAE increased by approximately 10\% and SSIM decreased significantly, while PSNR and TSD showed no significant differences. Notably, LUCID with in-house CT pretraining had higher MAE than the CNN prior in all 32 scans, suggesting that limited in-house CT data may provide insufficient anatomical priors for effective refinement, leading to degraded accuracy under limited paired supervision.

\subsubsection{Sampling Parameters} 

\begin{figure}[t]
    \centering
    \includegraphics[width=\linewidth]{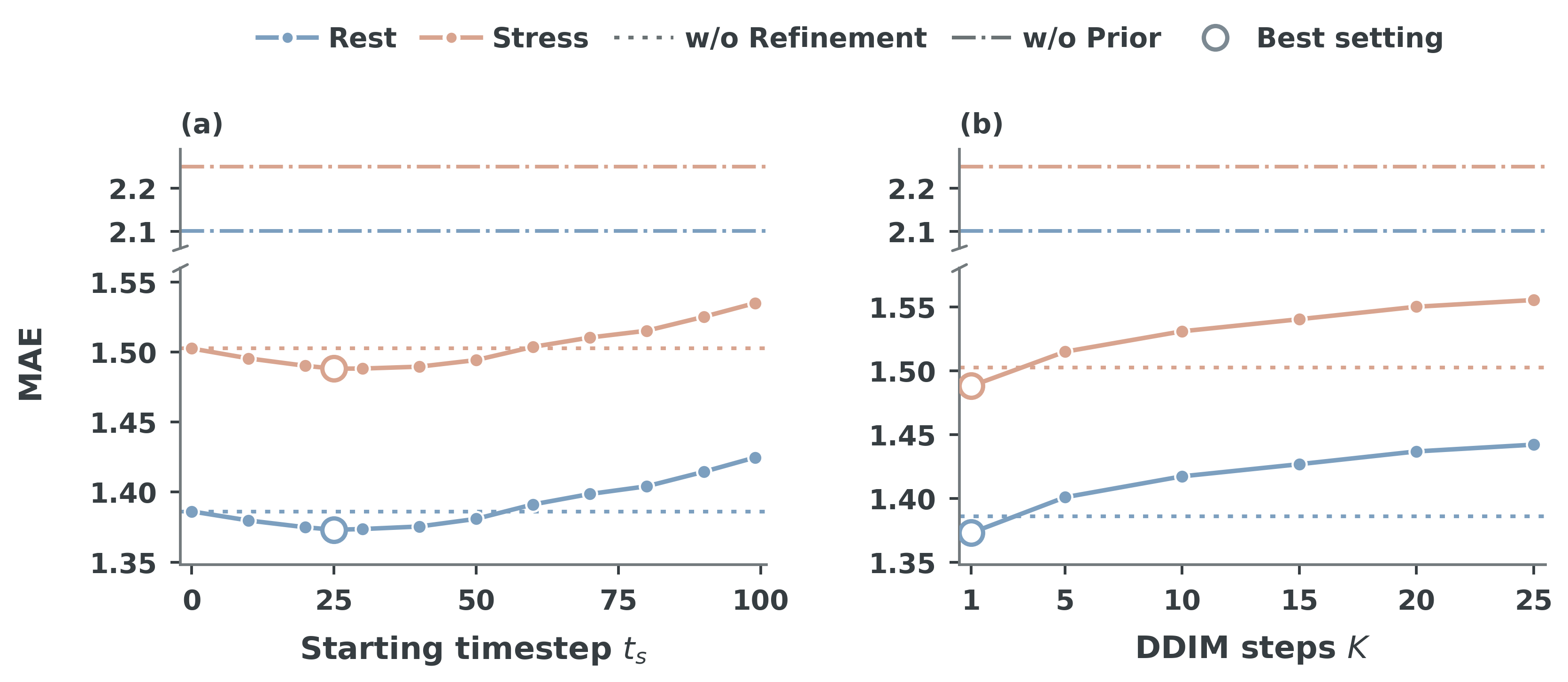}
    \caption{Sensitivity of LUCID to the sampling parameters. MAE is reported in units of $10^{-3}\,\mathrm{cm}^{-1}$.}
    \label{fig:ts_k}
\end{figure}

Fig.~\ref{fig:ts_k} shows the sensitivity of LUCID to the starting timestep $t_s$ and the number of DDIM~\cite{song2021denoising} sampling steps $K$. With $K=1$, MAE first decreased and then increased as $t_s$ increased, reaching its minimum at $t_s=25$. Setting $t_s=0$ corresponds to w/o Refinement. At $t_s\geq60$, MAE exceeded that of w/o Refinement but remained lower than w/o Prior. With $t_s=25$, increasing the sampling steps did not improve accuracy, and MAE exceeded that of w/o Refinement at $K=5$. These results show that light, low-noise diffusion refinement achieved the best accuracy, with $t_s=25$ and $K=1$ providing the lowest MAE at minimal sampling cost

\section{Discussion}

This study showed that LUCID enables accurate CT-free $\mu$-map synthesis for dynamic cardiac $^{82}$Rb PET. Compared with CNN-, GAN-, and diffusion-based methods, LUCID achieved the best synthesis accuracy in both rest and stress scans with competitive inter-frame consistency. The ablations indicated that this performance did not arise from stronger generative modeling alone, but from the combination of patient-specific CNN initialization, diffusion refinement with a large-scale CT prior, and the use of multiple dynamic frames.

The ablation studies distinguished the roles of CNN initialization and diffusion refinement. Removing the CNN prior increased MAE by approximately 50\%, indicating that starting from a regression prediction better preserves patient-specific attenuation information than generating from pure noise, consistent with prior-initialized diffusion methods~\cite{chen20252,zhou2024cmdm}. Removing refinement caused only a small decrease in accuracy but increased TSD in all 32 scans, indicating that refinement mainly improves inter-frame consistency.

Large-scale CT pretraining determined whether refinement improved accuracy. Compared with the 71 paired training patients, CT-RATE~\cite{hamamci2026generalist} provided a wider range of thoracic anatomical variation, tissue boundaries, and structural relationships. Even without the CNN prior, it achieved lower MAE than DDPM~\cite{ho2020ddpm} and BBDM~\cite{li2023bbdm}. This advantage disappeared with in-house CT pretraining, which yielded comparable performance to these baselines, indicating that the benefit came from large-scale pretraining rather than the ControlNet~\cite{zhang2023controlnet} architecture alone. For the full LUCID model, refinement with the in-house pretrained backbone reduced TSD in all 32 scans, but increased MAE relative to the CNN prior in every scan. Thus, refinement improved inter-frame consistency with both backbones, while accuracy improved only with large-scale CT pretraining.

The sampling analysis showed that diffusion refinement should remain light. Increasing the starting timestep or the number of sampling steps can raise MAE, sometimes making refinement worse than the initial CNN prediction, consistent with findings in iterative image restoration~\cite{delbracio2023inversion}. This behavior can be explained by the perception--distortion trade-off~\cite{blau2018dp}. Under mean squared error, the posterior mean minimizes distortion by averaging possible reconstructions, often producing smoother images. Improving perceptual realism requires moving away from this estimate, potentially increasing distortion. This agrees with our qualitative observations: regression methods produced smoother boundaries but lower errors, whereas generative baselines showed richer textures but higher errors. PMRF~\cite{ohayon2025posteriormean} further provides a theoretical basis for the importance of initialization, showing that optimal transport from the posterior mean can minimize distortion under perceptual constraints. Our starting timestep results support a similar principle. At a starting timestep of zero, refinement becomes an identity mapping, equivalent to w/o Refinement. Beyond the optimal low-noise range, stronger perturbations may weaken patient-specific information in the CNN prediction and increase MAE. The best accuracy at a low starting timestep indicates that refining a low-error CNN prediction by correcting residual errors is more effective. For $\mu$-map synthesis, accurate attenuation coefficients are more important than additional visual detail, making light refinement better suited to this task.

Beyond diffusion refinement, LUCID also benefits from multi-frame inference. Compared with static PET, individual dynamic PET frames have shorter acquisition times and higher noise levels, making frame-wise predictions less stable and potentially causing the model to mistake noise for anatomical structure. Although frames differ in tracer distribution, count level, and noise, they come from the same patient and should ideally share the same $\mu$-map. Since PET counts follow Poisson statistics~\cite{shepp1982maximum}, each frame contains a different noise realization. Averaging frame-wise predictions can therefore be viewed as an approximate Monte Carlo estimate~\cite{metropolis1949monte,wang2019aleatoric} of the expected prediction under input perturbations. When frame-wise errors are not fully correlated, random errors can partially cancel through averaging. Consistent with this explanation, single-frame inference yielded higher MAE in both rest and stress scans and lower PSNR in all scans, while input and output averaging achieved similar accuracy. However, systematic errors shared across frames cannot be removed by averaging.

Despite these findings, this study has several limitations. First, evaluation focused on $\mu$-map accuracy and inter-frame consistency. Whether these improvements translate into more accurate AC PET, relative perfusion~\cite{jorgensen2026attenuation}, and MBF quantification~\cite{sun2023transfer} remains to be evaluated. Since MBF estimation relies on early dynamic frames~\cite{murthy2018clinical}, the impact of $\mu$-maps synthesized from frames 19--27 on full kinetic analysis requires further investigation. Second, a single CT-derived $\mu$-map served as the reference for all dynamic frames, although CT and PET may capture different respiratory states. Strictly speaking, no reference $\mu$-map perfectly matches every dynamic frame. Respiratory mismatch and patient motion may contribute to residual errors near the diaphragm and hepatic dome. Motion correction~\cite{guo2024tai}, deformable registration~\cite{schaefferkoetter2023deep}, and explicit motion modeling~\cite{elhamiasl2025joint} may help reduce these errors. Third, refinement strength and frame weights are fixed. Light refinement may not fully correct large structural errors in the CNN initialization, while equal-weight averaging ignores differences in frame duration, count level, and noise. Adaptive refinement and frame weighting based on prediction uncertainty~\cite{NIPS2017_2650d608} or count statistics may improve robustness. Finally, the paired PET/CT data came from a single center, scanner system, and $^{82}$Rb tracer, with only 16 patients in the test set. Performance was also slightly lower for stress than rest scans. External validation across centers, scanners, patient populations, and tracers is needed to assess generalizability. Multi-institutional approaches, including federated learning~\cite{zhou2022federated} and multi-domain training~\cite{hajianfar2026reconstruction}, may further improve robustness.

\section{Conclusion}

This study presented LUCID, a CT-free $\mu$-map synthesis method for dynamic cardiac $^{82}$Rb PET. It starts from a CNN prediction, applies light diffusion refinement using a backbone pretrained on large-scale CT data, and averages frame-wise predictions. Evaluated on 102 patients, LUCID achieved the highest quantitative accuracy in both rest and stress scans compared with CNN-, GAN-, and diffusion-based methods, while maintaining competitive inter-frame consistency. Ablation studies showed that CNN initialization provides the foundation for accuracy, diffusion refinement improves inter-frame consistency but further improves accuracy only with large-scale CT pretraining, and multi-frame aggregation improves the final $\mu$-map accuracy. These findings support combining patient-specific predictions with large-scale CT anatomical priors for CT-free attenuation correction in dynamic PET. Future work will evaluate its impact on attenuation-corrected PET images and myocardial blood flow quantification.

\section*{Acknowledgments}

All authors declare that they have no known conflicts of interest in terms of competing financial interests or personal relationships that could have an influence or are relevant to the work reported in this paper.

\bibliographystyle{IEEEtran}
\bibliography{references}

\end{document}